\documentclass[runningheads]{llncs}

\usepackage{eccv}

\usepackage{eccvabbrv}

\usepackage{graphicx}
\usepackage{booktabs}
\usepackage[dvipsnames]{xcolor, colortbl}
\usepackage{graphicx}
\usepackage{booktabs}
\usepackage{booktabs} 
\usepackage{multirow}
\usepackage{caption} 
\usepackage{adjustbox}

\usepackage[accsupp]{axessibility}  

\usepackage{hyperref}

\usepackage{orcidlink}

\begin{document}

\title{Logit Disagreement: OoD Detection with Bayesian Neural Networks} 

\titlerunning{Logit Disagreement: OoD Detection with BNNs}

\author{Kevin Raina\inst{1}\orcidlink{0000-0002-6240-9675}}

\authorrunning{K.Raina}

\institute{Department of Mathematics and Statistics, University of Ottawa, ON, K1N 6N5, Canada \\
\email{krain033@uottawa.ca}}

\maketitle

\begin{abstract}
 Bayesian neural networks (BNNs), which estimate the full posterior distribution over model parameters, are well-known for their role in uncertainty quantification and its promising application in out-of-distribution detection (OoD). Amongst other uncertainty measures, BNNs provide a state-of-the art estimation of predictive entropy (total uncertainty) which can be decomposed as the sum of mutual information and expected entropy. In the context of OoD detection the estimation of predictive uncertainty in the form of the predictive entropy score confounds aleatoric and epistemic uncertainty, the latter being hypothesized to be high for OoD points. Despite these justifications, the mutual information score has been shown to perform worse than predictive entropy. Taking inspiration from Bayesian variational autoencoder (BVAE) literature, this work proposes to measure the disagreement between a corrected version of the pre-softmax quantities, otherwise known as logits, as an estimate of epistemic uncertainty for Bayesian NNs under mean field variational inference. The three proposed epistemic uncertainty scores demonstrate marked improvements over mutual information on a range of OoD experiments, with equal performance otherwise. Moreover, the epistemic uncertainty scores perform on par with the Bayesian benchmark predictive entropy on a range of MNIST and CIFAR10 experiments.

\end{abstract}

\section{Introduction}
\label{sec:intro}

Deep neural networks (DNNs) have shown strong predictive performance on a variety of machine learning tasks
such as computer vision \cite{krizhevsky2009learning,simonyan2014very}, natural language processing \cite{mikolov2010recurrent, mikolov2013efficient}, speech recognition \cite{senior2012deep, hannun2014deep} and bio-informatics \cite{alipanahi2015predicting}. Despite their
success, DNNs can erroneously produce confident predictions on inputs drawn from a different distribution than the training data, otherwise known as out-of-distribution (OoD) inputs \cite{nguyen2015deep, hendrycks2016baseline}. The detection of OoD inputs is crucial in safety-critical applications with harmful, offensive, or fatal consequences \cite{amodei2016concrete}. An added challenge is dealing with noisy or ambiguous datasets, which also appear in safety-critical applications \cite{huang2020autonomous,kamnitsas2016deepmedic}. It is clear that there is a need to equip DNNs with uncertainty awareness to prevent erroneous usage and signal human intervention when needed. 

Bayesian neural networks (BNNs) offer a principled approach for robust uncertainty quantification in deep learning \cite{neal2012bayesian, mackay1992bayesian, gal2016uncertainty} with inherent probabilistic interpretations. While exact Bayesian inference is often computationally intractable, several implementations of posterior inference have been proposed 
\cite{blundell2015weight, maddox2019simple, chen2014stochastic} enabling practical application.  In result, the Bayesian NN literature has grown with the proposal and adaptation of many uncertainty measures \cite{kendall2017uncertainties, kwon2022uncertainty, feinman2017detecting, wang2021bayesian}. 
Amongst many other uncertainty measures, predictive entropy (PE) and mutual information (MI) are recurring scores studied in the context of OoD detection \cite{lakshminarayanan2017simple, rawat2017adversarial, krishnan2020specifying, smith2018understanding, kirsch2021PEharmful}. 

\begin{figure}[h]
\label{Mukhoti}
\begin{center}
\includegraphics[scale=0.32]{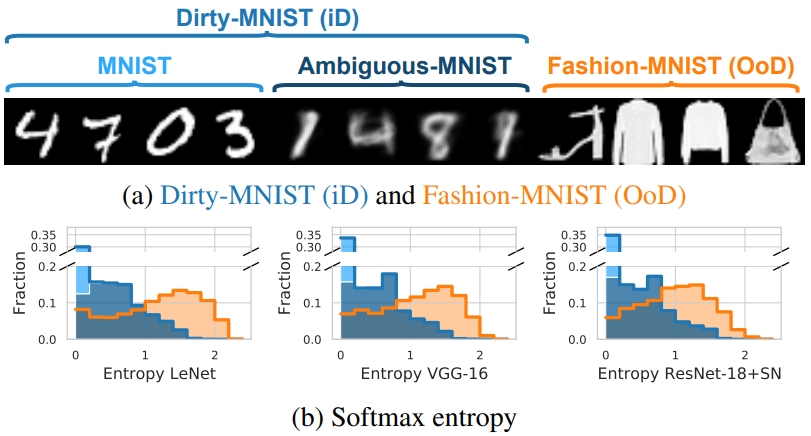}
\end{center}
\caption{(a): The DirtyMNIST dataset, comprising of classic MNIST \cite{deng2012mnist} and 60K samples of generated ambiguous digits (AmbiguousMNIST) with FashionMNIST \cite{xiao2017fashion} as OoD. (b): Entropy of DNNs, which capture aleatoric uncertainty, can not efficiently separate DirtyMNIST from FashionMNIST. Figure derived with permission of the authors \cite{mukhoti2023deep}.}
\label{fig:Mukhoti}

\end{figure}

In review of the dichotomy of uncertainty present in Bayesian modelling, there are two major sources of uncertainty: epistemic and aleatoric \cite{kendall2017uncertainties, der2009aleatory}. Epistemic uncertainty accounts for uncertainty in the model parameters which occurs due to lack of similar training data and is reducible with the incorporation of more data. Logically, OoD data are hypothesized to have high epistemic uncertainty as such inputs should not resemble the training data. Aleatoric uncertainty is uncertainty due to inherent noise and stochasticity in the observations and is high when presented with ambiguous or noisy inputs. Aleatoric uncertainty is irreducible, as we can not change the structure of an input with additional data. For instance, in Figure \ref{fig:Mukhoti} (a) the AmbiguousMNIST part of the DirtyMNIST dataset \cite{mukhoti2023deep} consists of hand-written digits that we may identify with one or more digits.

From the framework of uncertainty modelling, predictive entropy captures both aleatoric and epistemic uncertainty, while mutual information is thought to capture epistemic uncertainty. The incorporation of both uncertainties in predictive entropy offers an explanation to why this uncertainty measure can yield poor OoD detection performance when the in-distribution (iD) dataset consists of noisy inputs \cite{kirsch2021PEharmful}. For instance, Figure \ref{fig:Mukhoti} (b) provides histograms of softmax entropy from DNNs trained on DirtyMNIST, which contains an equal representation of the  dataset MNIST and AmbiguousMNIST. Mutual information, however, does not incorporate aleatoric uncertainty and may be more robust on such datasets. Crucially, we also do not have control over the ambiguity of an input, but its familiarity in the dataset. This rationale suggests another advantage of epistemic uncertainty scores such as mutual information in OoD detection. 

Despite theoretical justifications, the mutual information is shown to yield promising but not optimal performance on curated datasets \cite{malinin2018predictive}. One possible explanation can be attributed to the miscalibration of softmax scores \cite{guo2017calibration}. Several uncertainty quantification methods for deep neural networks have bypassed this issue by basing their scores on earlier DNN layer features, and have demonstrated improved OoD detection capabilities \cite{postels2020hidden, mukhoti2023deep, sun2022out, lee2018simple}. 
In this work, easy-to-implement and model-agnostic epistemic uncertainty scores based off measuring the disagreement in the maximum logit value across posterior samples are proposed and evaluated for BNNs. In particular the disagreement scores are taken from Bayesian variational autoencoder (BVAE) literature, wherein they are applied to marginal likelihoods instead \cite{glazunov2022bayesian, daxberger2020bayesian}.

\section{Background}\label{sect:background}

\paragraph{Bayesian Deep Learning.} The Bayesian approach is to treat the weights of a deep neural
network as random variables \cite{neal2012bayesian,mackay1992bayesian}. By assuming a prior distribution $p(\omega)$ on the model weights,  we can infer a posterior distribution $p(\omega|\mathcal{D})$ from a given dataset $\mathcal{D} = \{(x_i,y_i)\}_{i=1}^{N}$ via variational inference \cite{blundell2015weight}, wherein there is a need to specify the variational approximating distribution $q_{\theta}(\omega)$ indexed by variational parameter $\theta$. Variational inference aims to find the closest distribution to the true posterior from the parametric family $Q = \{q_{\theta}(\omega); \theta \in \Theta \}$ where closeness is often measured by Kullback-Leibler (KL) divergence. Consequently, we aim to minimize the objective function:

\begin{equation}
\begin{aligned}
\mathcal{F}(\mathcal{D},\omega) & =  \int q_{\theta}(\omega)\log\frac{q_{\theta}(\omega)} {p(\omega)p(\mathcal{D}|\omega)}d\omega \\
& = \mathbf{KL}[q_{\theta}(\omega)||p(\omega)] - \mathbb{E}_{q_{\theta}(\omega)}[\log p(\mathcal{D}|\omega)]
\label{eq:1}
\end{aligned}
\end{equation}

\noindent which can be approximated in practice by employing the reparametrization trick and Monte Carlo sampling from the variational distribution.
In the classification setting, each posterior sample $\omega_i \sim p(w|\mathcal{D})$ and input $x$ is associated with a categorical distribution $p(y|x, \omega)$ over the labels $y \in \mathcal{Y}$, \ie, an individual model from an infinite ensemble. The predictive distribution is obtained by marginalizing out the model weights over the posterior distribution: $p(y|x,\mathcal{D}) =  \mathbb{E}_{p(\omega|\mathcal{D})}[p(y|x,\omega)]$. In variational inference, expectations of a similar form are effectively approximated by sampling from the variational distribution and then using the Monte Carlo estimator

\begin{equation}
\begin{aligned}
\mathbb{E}_{p(\omega|\mathcal{D})}[f(\omega,x)] & = \int p(\omega|\mathcal{D})f(\omega,x)d\omega\\
& \approx \int q_{\hat{\theta}}(\omega)f(\omega,x)d\omega\\
& \approx \sum_{i=1}^{T} f(\omega_i,x), \omega_{1,...,T}\sim q_{\hat{\theta}}(\omega).\\
\label{eq:2}
\end{aligned}
\end{equation}

\paragraph{Uncertainty Quantification.} Within the Bayesian framework, a well-known decomposition is that of the predictive uncertainty, which itself quantifies total uncertainty. Following Smith and Gal's \cite{smith2018understanding}  terminology, the predictive entropy can be represented as the sum of mutual information (epistemic component) and expected entropy (aleatoric component), \ie, $\mathbb{H}[p(y|x,\mathcal{D})] = \mathbb{I}[y,\omega|x,\mathcal{D}] + \mathbb{E}_{p(\omega|\mathcal{D})}[\mathbb{H}[p(y|x,\omega)]]$, and these quantities can be approximated by (\ref{eq:2}).  The mutual information score can be expressed to intuitively quantify the expected information gain in the posterior distribution over labels, with the addition of an observable point $(x,y)$ augmented into the training set, \ie, if $\mathcal{D^*} = \mathcal{D} \cup \{(x,y)\}$, then

\begin{equation}
    \mathbb{I}[y,\omega|x,\mathcal{D}] = \mathbb{E}_{p(y|x,\mathcal{D})}[\mathbf{KL}[p(\omega|\mathcal{D^*})||p(\omega|\mathcal{D})]].
    \label{eq:3}
\end{equation}


\section{Methodology}


\paragraph{Intuition:} Following Bayesian VAE methods \cite{daxberger2020bayesian, glazunov2022bayesian}  that measure the disagreement in marginal likelihoods $\{ p(x|\omega_i)\}_{i=1}^{M}$ from a set $\Omega = \{ \omega_i\}_{i=1}^{M}$ of model parameter samples $\omega_i \sim q_{\hat{\theta}}(\omega)$ , the aim is to measure epistemic uncertainty in Bayesian NNs and apply it to OoD detection by measuring variation or disagreement in the conditional likelihoods $\{ p(\hat{y}|x,\omega_i) \}_{i=1}^{M}$, where $\hat{y}$ is the predicted label selected from the predictive distribution. Intuitively, if the models $\{ \omega_i\}_{i=1}^{M}$ agree on how likely/probable the predicted label y is associated with an input x, then x is in favor of being in-distribution. However, if the models $\{ \omega_i\}_{i=1}^{M}$ disagree on the likelihood/probability of associating the predicted label y with an input x, then x is in favor of being OoD.

\paragraph{Disagreement and Model Informativeness:} To see a stronger connection between model disagreement and epistemic uncertainty, reconsider the initial Bayesian NN setting of inferring a posterior distribution $p(\omega|\mathcal{D})$ from a likelihood $p(\mathcal{D}|\omega)$ and prior $p(\omega)$:

\begin{equation}
    p(\omega|\mathcal{D}) = \frac{p(\mathcal{D}|\omega)}{p(\mathcal{D})}p(\omega).
    \label{eq:4}
\end{equation}

\noindent Another distribution of interest is the augmented posterior, which is the posterior distribution updated with the additional observation $(x,y)$. In this context, $\mathcal{D}^* = \mathcal{D} \cup \{(x,y) \}$ and the augmented posterior can be inferred as

\begin{equation}
    p(\omega|\mathcal{D}^*) = \frac{p(y|x,\omega)}{p(y|x,\mathcal{D})}p(\omega|\mathcal{D}),
    \label{eq:5}
\end{equation}

\noindent where it is readily observable that the change in the posterior distribution at model $\omega$ from augmenting the training data with observation $(x,y)$ depends on the ratio $\frac{p(y|x,\omega)}{p(y|x,\mathcal{D})}$, also referred to as the normalized likelihood. Applying (\ref{eq:2}) to estimate the predictive distribution results in

\begin{equation}
    \frac{p(y|x,\omega)}{p(y|x,\mathcal{D})} \approx \frac{p(y|x,\omega)}{\frac{1}{M} \sum_{\omega \in \Omega} p(y|x,\omega)} = M\eta_{\omega}; \hspace{1em} \text{where} \hspace{1em} \eta_{\omega} = \frac{p(y|x,\omega)}{\sum_{\omega \in \Omega} p(y|x,\omega)}.
    \label{eq:6}
\end{equation}

\noindent From the approximation (\ref{eq:6}), $\frac{p(y|x,\omega)}{p(y|x,\mathcal{D})} \approx 1$ when $\eta_w = \frac{1}{M}$, or there is no change in the posterior from augmenting observation $(x,y)$ if all models agree on its likelihood. Conversely, more disagreement amongst model likelihoods, will lead to more change in the posterior distribution. Likewise, we can derive a similar structure when specifically considering the KL Divergence between $p(\omega|\mathcal{D})$ and $p(\omega|\mathcal{D^*})$. In particular:

\begin{equation}
\begin{aligned}
 \mathbf{KL}[p(\omega|\mathcal{D})||p(\omega|\mathcal{D}^*)] & = \mathbb{E}_{p(\omega|\mathcal{D})} \biggl[ \log\frac{p(\omega|\mathcal{D})}{p(\omega|\mathcal{D}^*)} \biggr] \\
& = -\mathbb{E}_{p(\omega|\mathcal{D})}\biggl[\log\frac{p(\omega|\mathcal{D^*})}{p(\omega|\mathcal{D})} \biggr]\\
& = -\mathbb{E}_{p(\omega|\mathcal{D})} \biggl[ \log\frac{p(y|x,\omega)}{p(y|x,\mathcal{D})}\biggr] \hspace{1em} (\ref{eq:5})\\
& \approx -\mathbb{E}_{p(\omega|\mathcal{D})} \biggl[\log M\eta_{\omega} \biggr] \hspace{1em} (\ref{eq:6})\\
& \approx -\frac{1}{M} \sum_{\omega \in \Omega} \log M\eta_{\omega} \hspace{1em}  (\ref{eq:2})
\end{aligned}
\label{eq:7}
\end{equation}

\noindent where again, setting $\eta_\omega = \frac{1}{M}$ (all models agree on the likelihood) results in zero KL divergence. Moreover, when one model dominates the likelihood scores, \ie, $\eta_\omega \approx 1$ for some $\omega$, then the disagreement is high and the KL Divergence will be maximized.  We observe model disagreement being related to informativeness and epistemic uncertainty. Notice that throughout the derivations, $y$ was assumed to be known, which is seldom the case in practice. To work around this, we can either use the predicted label, or take an expectation over the predictive distribution. The former is considered in this work, whereas the latter is undertaken by the baseline score mutual information.

\paragraph{Logit Proxy:} Despite the justifications of disagreement scores, it is well-known that the softmax probabilities are often miscalibrated, and produce consistently overconfident predictions on certain OoD samples \cite{guo2017calibration,nguyen2015deep}. Recent works in DNNs have shown promising structure in the pre-softmax outputs, otherwise known as logits, for OoD detection \cite{hendrycks2019scaling,wei2022mitigating}. Specifically the maximum logit method (MLS) \cite{hendrycks2019scaling}, equivalently to apply the logit value associated with the class of the predicted label, has outperformed the traditional maximum softmax probability (MSP) \cite{hendrycks2016baseline} in OoD detection on multi-class datasets, despite bearing similarities. From a Bayesian perspective, measuring disagreement between maximum logit values across posterior samples  $\{ z(\hat{y},x,\omega_i) \}_{i=1}^{M}$ in place of softmax likelihoods $\{ p(\hat{y}|x,\omega_i) \}_{i=1}^{M}$ may prove to be a relatively simple yet improved measure of epistemic uncertainty with improved OoD detection abilities. Moreover, this substitution requires no further training, and is applicable to pre-trained Bayesian neural networks, with inherent advantages of being easy-to-use and model-agnostic. Many existing epistemic uncertainty scores from Bayesian VAEs operate exclusively on non-negative  likelihood values. In order to allow such scores to operate on the maximum logit, a logit truncation is required: 


\begin{figure}[h!]
\begin{center}
\includegraphics[scale=0.45]{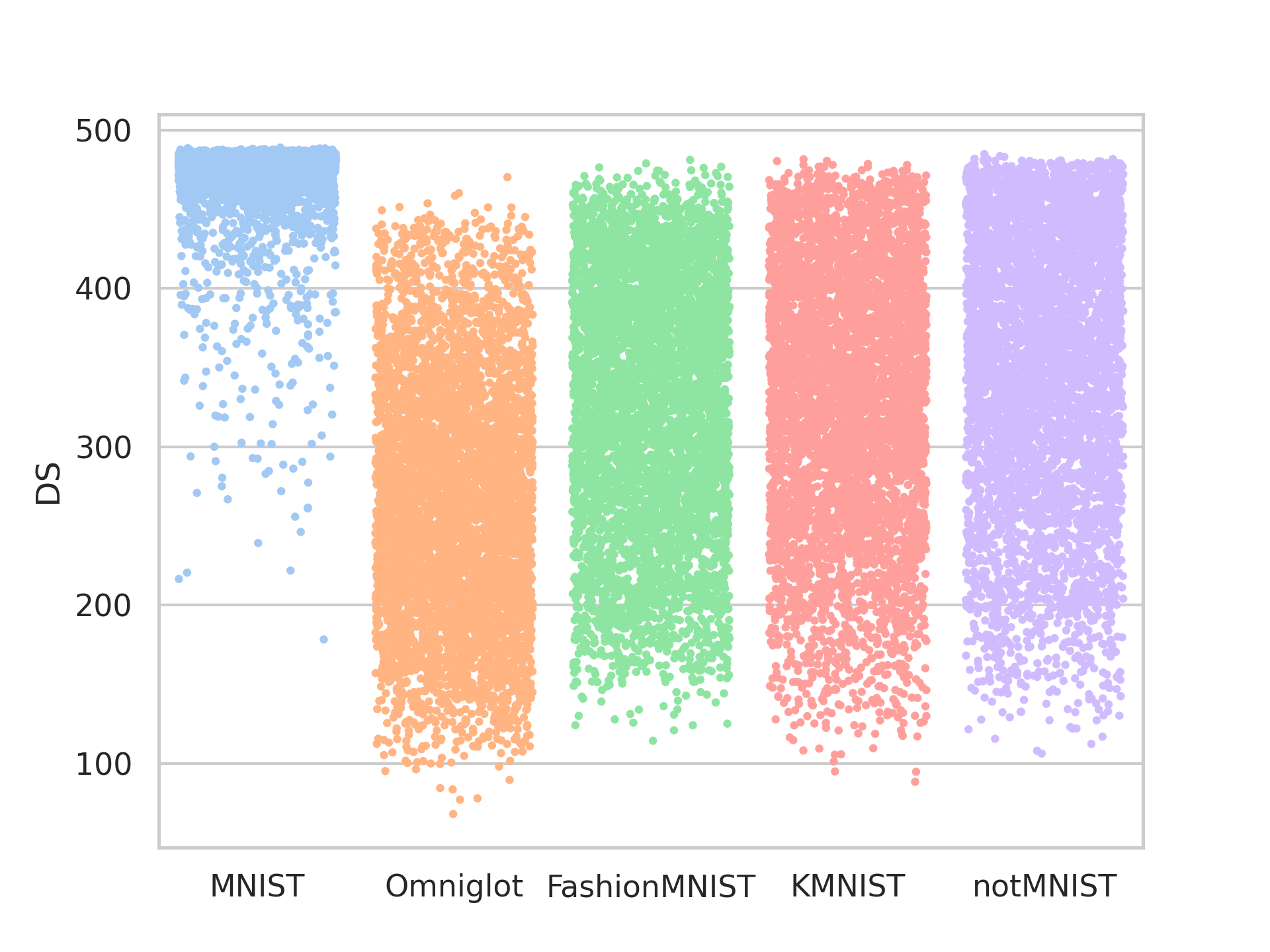}
\end{center}
\begin{center}
\includegraphics[scale=0.45]{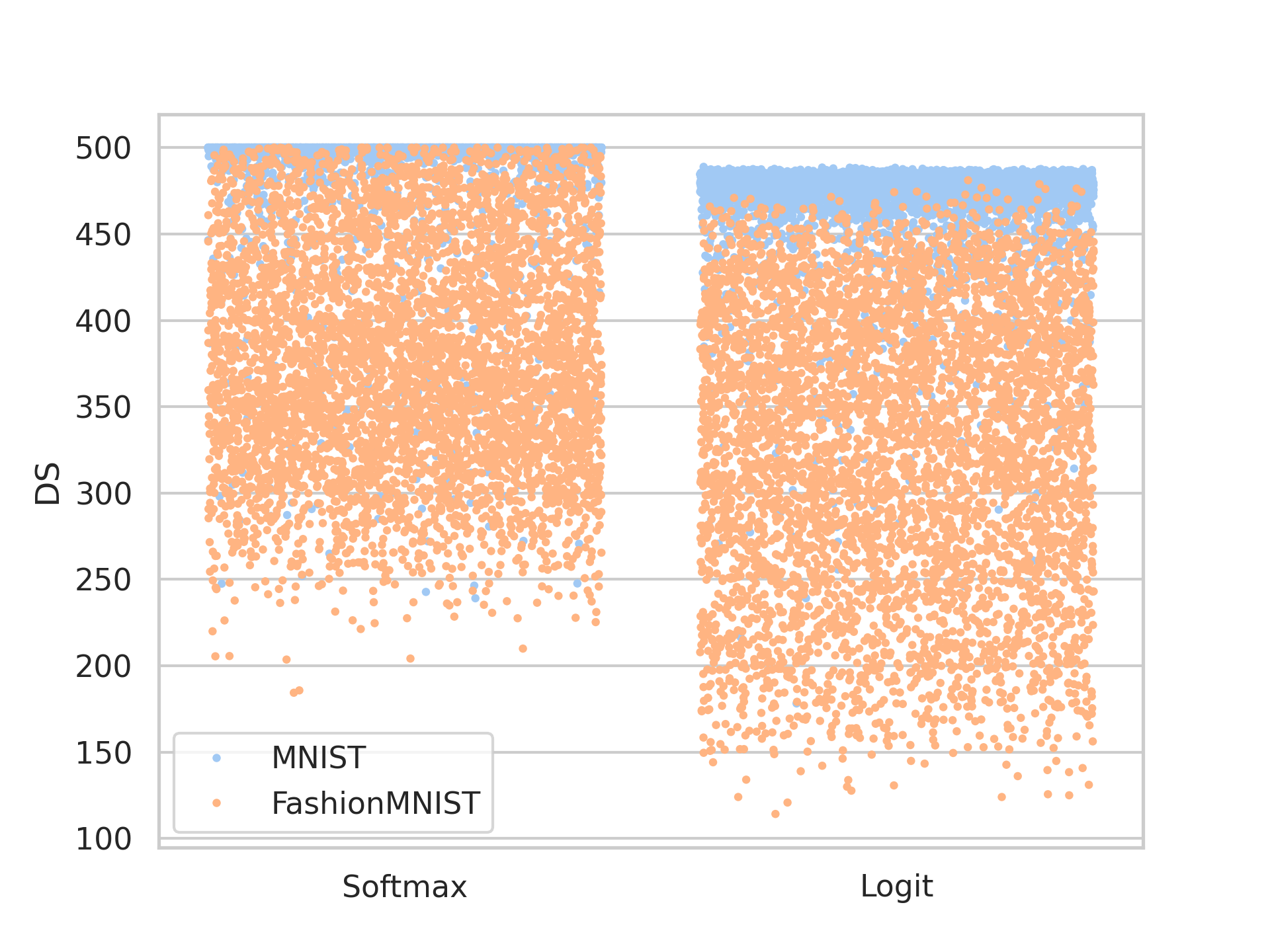}
\end{center}
\caption{Epistemic uncertainty for OoD using logit disagreement. Top: Strip plot of the disagreement score (DS) using the truncated logit score computed from a Bayesian LeNet-5 \cite{lecun1998gradient} trained on MNIST categorized by dataset. Bottom: A comparison between the disagreement score applied to the proposed logit proxy and the softmax  likelihood for MNIST vs FashionMNIST.}
\label{fig2:strip_plot}
\end{figure}

\begin{equation}
  z^{*}(\hat{y},x, \omega)=\begin{cases}
    z(\hat{y},x, \omega) & \text{if $z(\hat{y},x, \omega) > 0$},\\
    \epsilon & \text{otherwise}.
  \end{cases}
  \label{eq:8}
\end{equation}

\noindent where $\epsilon$ is a small positive constant (\eg $1\times 10^{-15}$). In the experiments it is important to note that virtually all of the structure in the max logit can be captured at non-negative values, as negative max logits can be considered to be outliers for both in-distribution and OoD samples. Our goal is now to measure disagreement amongst max logit values $\{z^{*}(\hat{y},x,\omega) \}_{i=1}^{M}$, corresponding to posterior samples $\{ \omega_i\}_{i=1}^{M}$ as a proxy to the softmax likelihoods.

Daxberger and Hernández-Lobato \cite{daxberger2020bayesian} suggests to calculate the disagreement score (DS) $D[\cdot]$ on marginal  likelihoods, which captures uncertainty of the models about the specific input. The DS is justified as the effective sample size (ESS) \cite{martino2017effective} in an importance sampling estimation framework, where $p(\omega|\mathcal{D})$ is the proposal distribution and $p(\omega|\mathcal{D^*})$ is the target distribution. Equivalently, DS quantifies the change in distribution between the proposal and target. By simply substituting the proposed logit proxy in place of the softmax likelihood, we obtain: 

\begin{equation}
\Tilde{D}_{\Omega}[x] = \frac{1}{\sum_{\omega \in \Omega} \Tilde{\eta}_{\omega}^2}; \hspace{1em} \text{with} \hspace{1em} \Tilde{\eta}_{\omega} = \frac{z^*(\hat{y}, x, \omega)}{\sum_{\omega \in \Omega} z^*(\hat{y}, x, \omega)},
\label{eq:9}
\end{equation}

\noindent
while the orginal softmax-based DS operates on the $\eta_{\omega}$ defined in
(\ref{eq:6}) instead. The max logits are first normalized and represented as weights $\Tilde{\eta}_{\omega}$ defined in (\ref{eq:9}), which can be interpreted as the probability that the sample was generated from model $\omega$. In the extreme case that all max logit values are identical, the weight values will follow a discrete uniform distribution. At the other extremity, one max logit value may dominate resulting in a degenerate distribution with all probability mass at a single  model. The disagreement score is the composition of first summing up the squares of the weights and then taking the reciprocal. If the score is large, it means that all models return close values of the logits. On the contrary, if the score is 1, then there is one model that dominates. Figure \ref{fig2:strip_plot} demonstrates the difference between applying the DS to the truncated logits and softmax scores in OoD detection on MNIST.

Additionally considered are two variants of measuring diagreement between marginal likelihoods proposed by Glazunov et al. \cite{glazunov2022bayesian} which are adapted to the max logits. Firstly, the weight entropy (WE) measures the Shannon information entropy of the normalized max logits $\Tilde{
\eta}_{\omega}$:

\begin{equation}
    \Tilde{\mathbb{H}}_{\Omega}[x] = - \sum_{\omega \in \Omega}  \Tilde{\eta}_{\omega} \log \Tilde{\eta}_{\omega}.
\end{equation}

\noindent 
Higher entropy will associate with more uniform plausibility across model parameters. Conversely, low entropy suggest lower uniformity in the choice of potential models, and higher disagreement in model possibilities. Secondly, sample standard deviation of the log-logits (Std of LLs) is considered:

\begin{equation}
    \Tilde{\Sigma}_{\Omega}[x] = \sqrt{ \frac{1}{M-1}\sum_{\omega \in \Omega}(\log z^*(\hat{y}, x, \omega) - \overline{\log z^*(\hat{y}, x, \omega)})^{2} }
\end{equation}

\noindent

\noindent
which is a classical estimator of variation. Naturally higher values lead to larger uncertainty arising from the posterior distribution.

\section{Experiments and Observations}\label{sect:experiments}

\paragraph{BCNN details.} The PyTorch Bayesian CNN implementation by Shridhar et al. \cite{shridhar2019comprehensive}, which efficiently applies the Bayes by Backprop method of Blundell et al. \cite{blundell2015weight} to CNNs 
is adapted for the experiments. For MNIST experiments, we follow Shridhar et al. and use the Softplus activation with $\beta=1$ on the LeNet-5 \cite{lecun1998gradient} CNN architecture. We place an isotropic Gaussian prior with $0$ mean and variance $0.01$ over the model weights, \ie, $p(\omega) = \mathcal{N}(0,0.01I)$. The variational posterior is taken as a diagonal Gaussian distribution with individual weights $\omega_i \sim \mathcal{N}(\omega_i|\mu_i, \sigma_i^2)$, where $\sigma_i = \log(1 + exp(\rho_i))$, and $\mu_i,\rho_i$ are initialized from $\mathcal{N}(0,0.01)$, $N(-5,0.01)$ respectively. Gradient descent is perfomed on the variational parameters $\theta = (\mu, \rho)$ via the Adam Optimizer with an initial learning rate $0.001$ on batches of size $256$ for a total of $200$ epochs. From mini-batches, the expected log- likelihood term from the variational objective is estimated with the stochastic gradient variational Bayes SGVB estimator\cite{kingma2015variational} and the analytical divergence term is multiplied by $\pi = 0.1$ for relative weighting. At each epoch, variational parameters are saved when the validation accuracy improves on a hold-out validation set.

In the CIFAR10 \cite{krizhevsky2009learning} experiments a Bayesian ResNet34 CNN is trained with informative priors. The MOPED method \cite{krishnan2020specifying} is applied with $\delta=0.01$ to establish the prior distribution of each weight by employing weights pre-trained on the ImageNet1K \cite{zhang2021understanding} dataset. At inference time, the diagonal standard deviations of the linear layer is scaled by a factor of $100$ to introduce higher variation from the posterior. Gradient descent is performed on the variational parameters $\theta = (\mu, \rho)$ via the Adam Optimizer with an initial learning rate $0.0005$ on batches of size $32$ for a total of $200$ epochs. Likewise, the expected log-likelihood term from the variational objective is estimated with the stochastic gradient variational Bayes SGVB estimator \cite{kingma2015variational} and the analytical divergence term is multiplied by $\pi = 0.1$ for relative weighting. Similar to MNIST training, variational parameters are saved when the validation accuracy improves on a hold-out validation set.

\begin{figure}
\begin{subfigure}{.50\textwidth}
  \centering
  \includegraphics[scale=0.25]{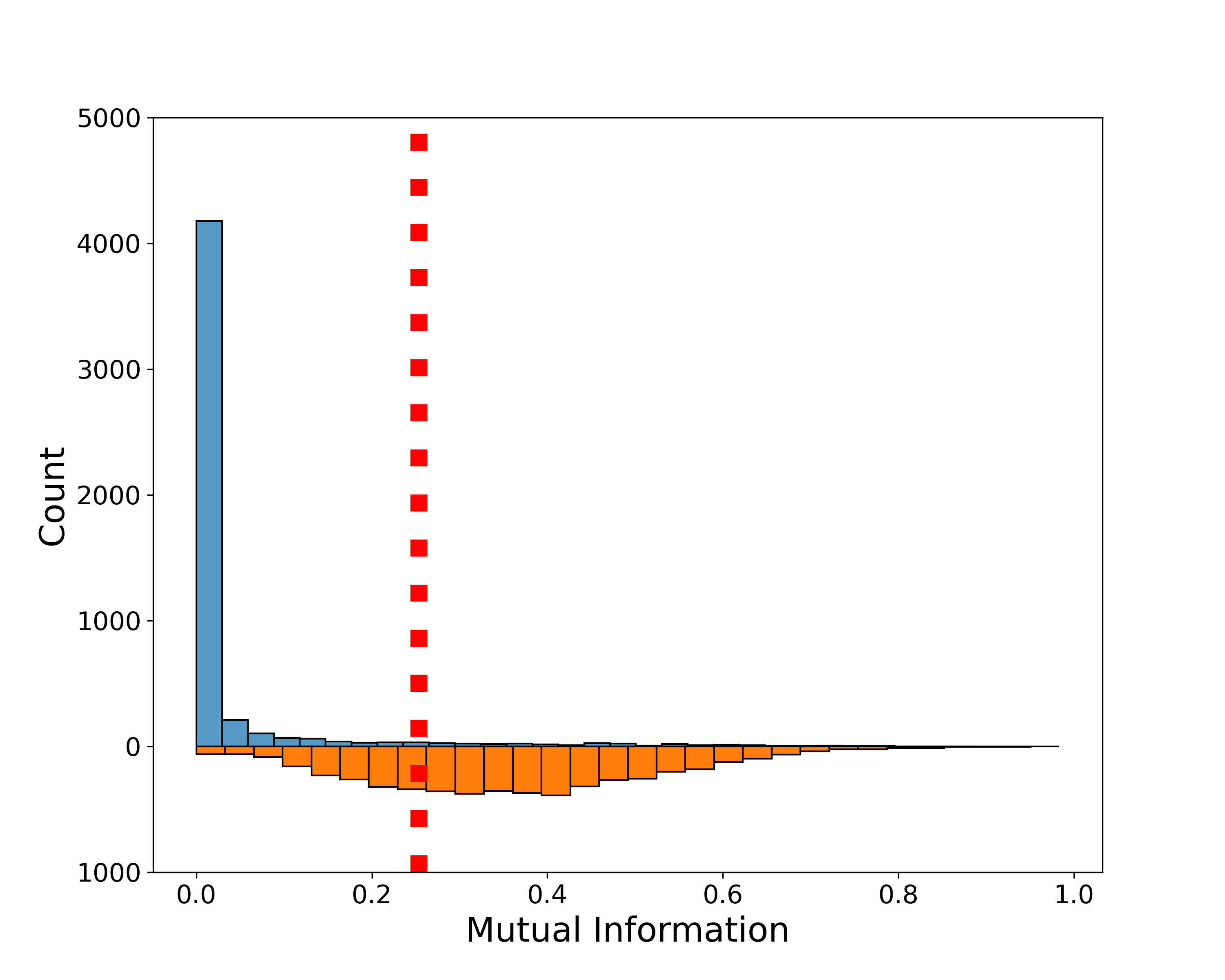}
\end{subfigure}
\begin{subfigure}{.50\textwidth}
  \includegraphics[scale=0.25]{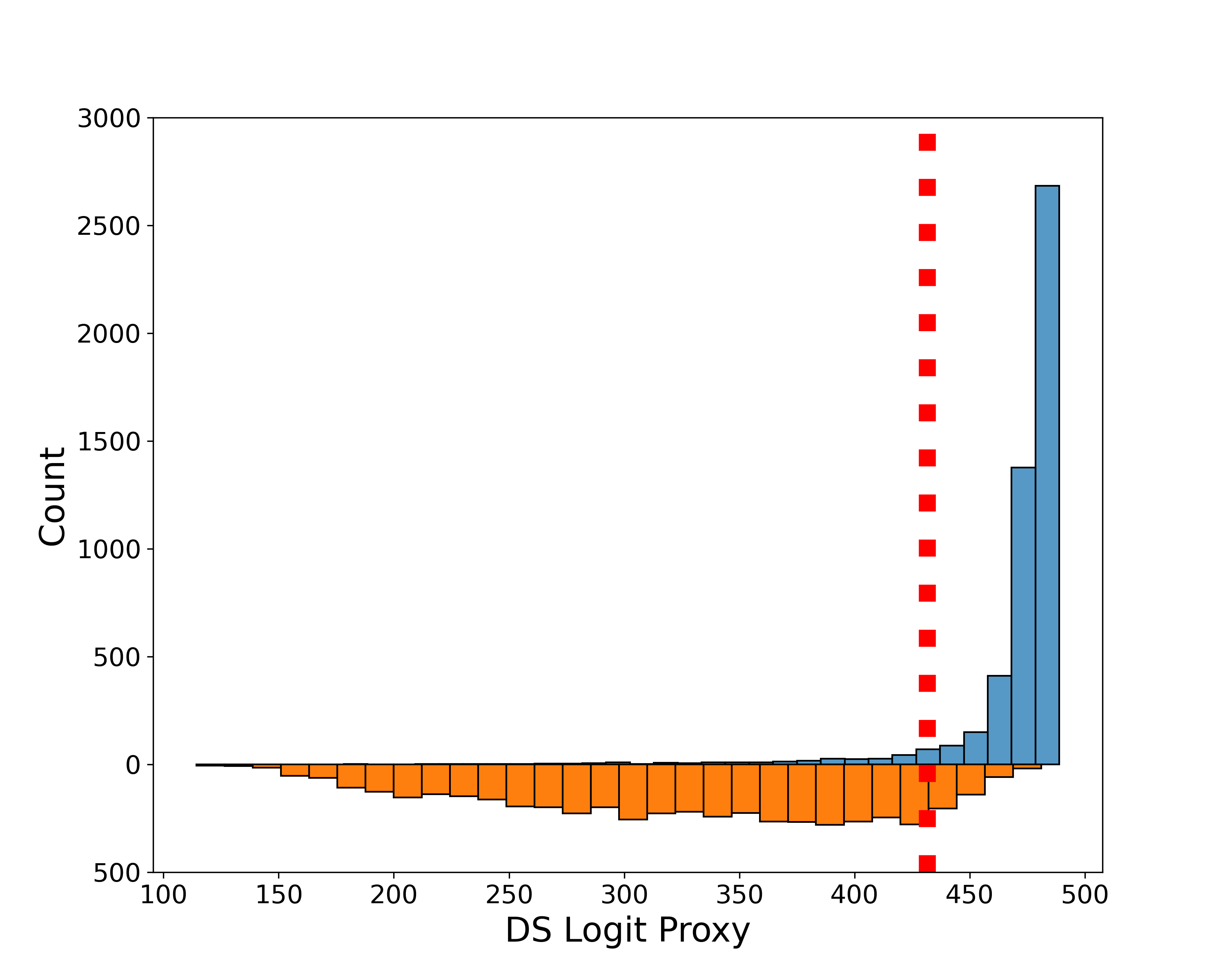}
\end{subfigure}
\caption{Histograms of select epistemic uncertainty scores for MNIST vs FashionMNIST, blue depicts in-distribution (iD) and orange - out-of-distribution (OoD). The red dotted line marks the $5$th or $95$th quantile of the iD scores. From left to right: MI and DS (logit proxy).}
\label{fig3:epistemic}
\end{figure}

\begin{figure}
\begin{subfigure}{.50\textwidth}
  \centering
  \includegraphics[scale=0.25]{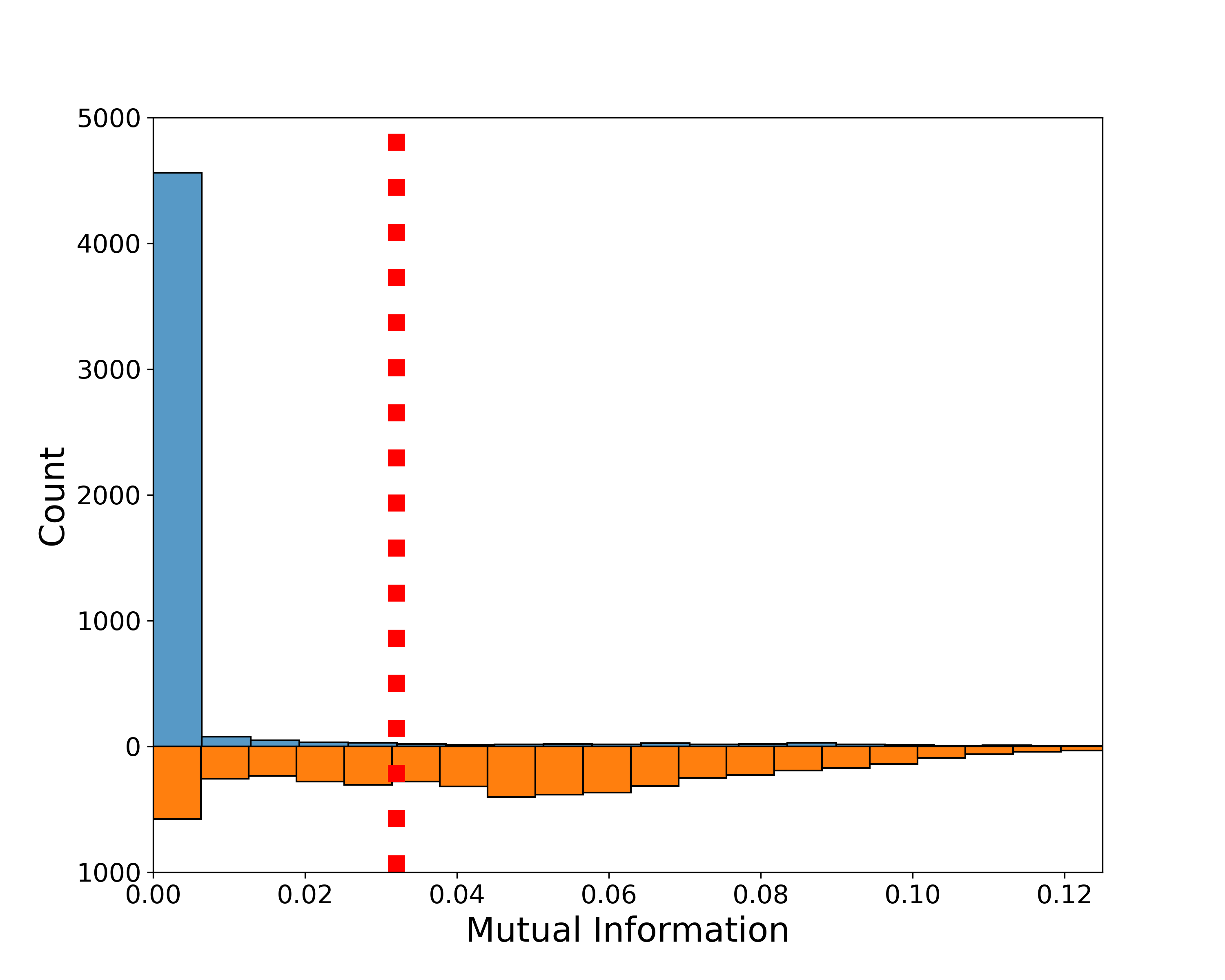}
\end{subfigure}
\begin{subfigure}{.50\textwidth}
  \includegraphics[scale=0.25]{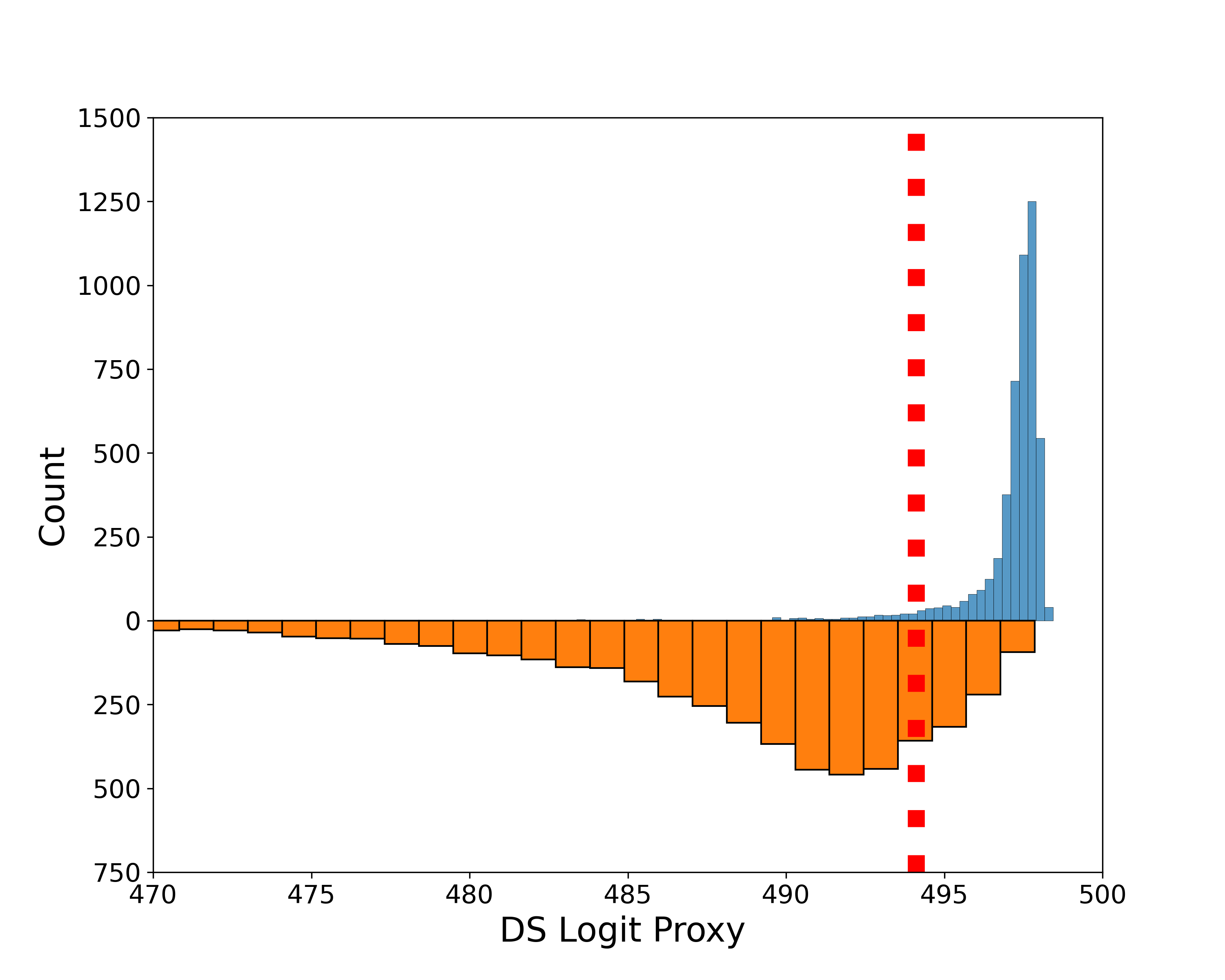}
\end{subfigure}
\caption{Histograms of select epistemic uncertainty scores for CIFAR10 vs SVHN, blue depicts in-distribution (iD) and orange - out-of-distribution (OoD). The red dotted line marks the $5$th or $95$th quantile of the iD score. From left to right: MI and DS (logit proxy).}
\label{fig4:predictive_uncertainty}
\end{figure}

\paragraph{Experimental setting.} Models are trained on the in-distribution dataset by randomly dividing the training set into an effective training set comprising $80$ percent of data and the remaining $20$ percent constituting the hold-out validation set. The model training and validation accuracies are reported in Table \ref{Table1}. In the first set of experiments, MNIST is the in-distribution with FashionMNIST \cite{xiao2017fashion} (fashion-product images), Omniglot \cite{lake2015human} (handwritten characters from 50 alphabets) , KuzushijiMNIST \cite{clanuwat2018deep} (kanji characters) and notMNIST \cite{bulatov2011notmnist} (various fonts of letters A through I) as OoD. For the second set of experiments CIFAR10 is the in-distribution with SVHN \cite{netzer2011reading}(cropped house number plates), CIFAR100 (superset of CIFAR10), Places365 \cite{zhou2017places} (scene recognition) and Textures \cite{cimpoi2014describing} (texture and pattern images) as OoD. Following Daxberger et al. \cite{daxberger2020bayesian} $5000$ in-distribution inputs are randomly sampled from the in-distribution set and $5000$ OoD inputs are randomly sampled from each of the OoD datasets. 
For evaluation the positive class is treated as OoD.
It is important to note that there is some disagreement in recent literature on which class to designate as positive; for example \cite{sun2022out, vernekar2019out}
define in-distribution samples to be positive, \ie, the opposite convention to the one presented.
For this reason, the false-negative rate in Tables \ref{Table1} and \ref{Table2} is comparable to the false-positive rates in \cite{vernekar2019out}.
The following metrics are computed: (1) The area under the ROC curve (AUROC↑), and (2) false-negative rate at $95$ percent true-negative rate (FNR95$\downarrow$). Sampling was generally carried out to preserve an equal representation of in-distribution and OoD labels. Notable instances are the Omniglot dataset, wherein a random selection of $250$ from the $1623$ labels were considered, and Places365 wherein images were randomly sampled. Nguyen et al. \cite{nguyen2022out} remarks that taking fewer samples choice provides a more realistic assumption as it is generally be unfeasible to observe OoD data of the same size, providing a motivation for the designated sampling. For the computation of all scores, $M=500$ models from the posterior distribution were sampled. Figures \ref{fig3:epistemic} and \ref{fig4:predictive_uncertainty} depict histograms of the max logit disagreement score along with the baseline mutual information on the MNIST vs FashionMNIST and CIFAR10 vs SVHN experiments respectively.

\begin{table*}[h!]
\centering
\begin{adjustbox}{max width=\textwidth}
\begin{tabular}{cccccccccc}
\toprule

\multicolumn{2}{c}{Dataset} & \multicolumn{4}{c}{AUROC $\uparrow$} & \multicolumn{4}{c}{FNR95$\downarrow$}\\
\midrule
\textbf{$\mathcal{D}_{in}$}  & \textbf{$\mathcal{D}_{out}$} & \textbf{MI} & \textbf{DS} & \textbf{Std of LLs} & \textbf{WE} & \textbf{MI} & \textbf{DS} & \textbf{Std of LLs} & \textbf{WE}\\

\midrule

 \multirow{1}{*}{} & FashionMNIST & $95.81$ & $98.14$ & $98.01$ & \cellcolor{lightgray}$\mathbf{98.16}$ & $28.58$ & \cellcolor{lightgray}$\mathbf{8.74}$ & $10.74$ & $9.40$  \\
  \multirow{1}{*}{MNIST}& Omniglot & $96.00$ & \cellcolor{lightgray}$\mathbf{99.39}$ & $99.26$ & $99.37$ & $18.54$ & \cellcolor{lightgray}$\mathbf{1.30}$ & $2.06$ & $1.60$\\
 \multirow{1}{*}{(99.0/98.5)}& KMNIST & $97.32$ & $97.80$ & $97.77$ & \cellcolor{lightgray}$\mathbf{97.81}$ & $14.98$ & \cellcolor{lightgray}$\mathbf{11.72}$ & $12.82$ & $11.94$ \\
 \multirow{1}{*}{}& notMNIST & $95.41$ & $95.89$ & \cellcolor{lightgray}$\mathbf{95.98}$ & $95.95$ & $24.76$ & $22.44$ & \cellcolor{lightgray}$\mathbf{22.38}$ & $22.68$ \\

\midrule

 \multirow{1}{*}{} & SVHN & $94.77$ & \cellcolor{lightgray}$\mathbf{96.84}$ & $96.83$ & $96.83$ & $33.58$ & $\cellcolor{lightgray}\mathbf{16.08}$ & $16.14$ & $16.10$ \\
  \multirow{1}{*}{CIFAR10}& CIFAR100  & \cellcolor{lightgray}$\mathbf{89.79}$ & $89.76$ & $89.76$ & $89.76$ & $53.50$ & \cellcolor{lightgray}$\mathbf{44.38}$ & $44.60$ & $44.46$ \\
     \multirow{1}{*}{(99.0/96.1)}& Places365 & $93.69$ & \cellcolor{lightgray}$\mathbf{95.28}$ & \cellcolor{lightgray}$\mathbf{95.28}$ & \cellcolor{lightgray}$\mathbf{95.28}$ & $41.90$ & \cellcolor{lightgray}$\mathbf{25.70}$ & \cellcolor{lightgray}$\mathbf{25.70}$ & \cellcolor{lightgray}$\mathbf{25.70}$ \\

 \multirow{1}{*}{}& Textures & $93.86$ & \cellcolor{lightgray}$\mathbf{95.02}$ & \cellcolor{lightgray} $\mathbf{95.02}$ & \cellcolor{lightgray}$\mathbf{95.02}$ & $43.80$ & \cellcolor{lightgray}$\mathbf{34.00}$ & \cellcolor{lightgray}$\mathbf{34.00}$ & $34.10$ \\

\bottomrule

\end{tabular}
\end{adjustbox}
\caption{Comparison of epistemic uncertainty scores.}
\label{Table1}
\end{table*}

\begin{table*}[h!]
\centering
\begin{adjustbox}{max width=\textwidth}
\begin{tabular}{cccccccccc}
\toprule

\multicolumn{2}{c}{Dataset} & \multicolumn{4}{c}{AUROC $\uparrow$} & \multicolumn{4}{c}{FNR95$\downarrow$}\\
\midrule
\textbf{$\mathcal{D}_{in}$}  & \textbf{$\mathcal{D}_{out}$} & \textbf{PE} & \textbf{DS} & \textbf{Std of LLs} & \textbf{WE} & \textbf{PE} & \textbf{DS} & \textbf{Std of LLs} & \textbf{WE}\\

\midrule

 \multirow{1}{*}{} & FashionMNIST & \cellcolor{lightgray}$\mathbf{98.77}$ & $98.14$ & $98.01$ & $98.16$ & \cellcolor{lightgray}$\mathbf{5.96}$ & $8.74$ & $10.74$ & $9.40$  \\
  \multirow{1}{*}{MNIST}& Omniglot & \cellcolor{lightgray}$\mathbf{99.82}$ & $99.39$ & $99.26$ & $99.37$ & \cellcolor{lightgray}$\mathbf{0.50}$ & $1.30$ & $2.06$ & $1.60$\\
 \multirow{1}{*}{(99.0/98.5)}& KMNIST & $97.29$ & $97.80$ & $97.77$ & \cellcolor{lightgray}$\mathbf{97.81}$ & $15.58$ & \cellcolor{lightgray}$\mathbf{11.72}$ & $12.82$ & $11.94$ \\
 \multirow{1}{*}{}& notMNIST & \cellcolor{lightgray}$\mathbf{96.00}$ & $95.89$ & $95.98$ & $95.95$ & \cellcolor{lightgray}$\mathbf{22.08}$ & $22.44$ & $22.38$ & $22.68$ \\

\midrule

 \multirow{1}{*}{} & SVHN & $96.65$ & \cellcolor{lightgray}$\mathbf{96.84}$ & $96.83$ & $96.83$ & $21.56$ & $\cellcolor{lightgray}\mathbf{16.08}$ & $16.14$ & $16.10$ \\
  \multirow{1}{*}{CIFAR10}& CIFAR100  & \cellcolor{lightgray}$\mathbf{90.30}$ & $89.76$ & $89.76$ & $89.76$ & $50.72$ & \cellcolor{lightgray}$\mathbf{44.38}$ & $44.60$ & $44.46$ \\
     \multirow{1}{*}{(99.0/96.1)}& Places365 & $94.51$ & \cellcolor{lightgray}$\mathbf{95.28}$ & \cellcolor{lightgray}$\mathbf{95.28}$ & \cellcolor{lightgray}$\mathbf{95.28}$ & $36.94$ & \cellcolor{lightgray}$\mathbf{25.70}$ & \cellcolor{lightgray}$\mathbf{25.70}$ & \cellcolor{lightgray}$\mathbf{25.70}$ \\

 \multirow{1}{*}{}& Textures & $94.90$ & \cellcolor{lightgray}$\mathbf{95.02}$ & \cellcolor{lightgray} $\mathbf{95.02}$ & \cellcolor{lightgray}$\mathbf{95.02}$ & $37.06$ & \cellcolor{lightgray}$\mathbf{34.00}$ & \cellcolor{lightgray}$\mathbf{34.00}$ & $34.10$ \\

\bottomrule

\end{tabular}
\end{adjustbox}
\caption{Comparison of the proposed scores with the Bayesian benchmark (predictive entropy).}
\label{Table2}
\end{table*}

\paragraph{Results.} Table \ref{Table1} shows the proposed methods compared to the standard Bayesian epistemic uncertainty score. On benchmark tasks involving MNIST, the methods demonstrates improved OoD detection. In particular, there are significant improvements in the false-negative rate at $95$ percent true-negative rate. Moreover, improvements in both AUROC and FNR95 are substantial for the far-OoD datasets Omniglot and FashionMNIST. The proposed methods always yielded the optimal performance in comparison to mutual information. Likewise on benchmark CIFAR10 tasks, the proposed methods show significant improvement in FNR95 across all experiments, and marked improvements in AUROC for SVHN, Places365 and Textures. In observation, the CIFAR100 experiment proved particularly difficult for all scores, but the proposed methods performed similar to mutual information in terms of AUROC.  In observation, mutual information appears to have a longer tail associated with its distribution contributing to worse OoD detection performance at extreme thresholds. Comparing the three proposed epistemic uncertainty scores to each other, note the disagreement score often performs the best in FNR95, but generally performs similarly to the standard deviation of log-logits and weight entropy.  Predictive entropy demonstrates decent performance, and outperforms mutual information on these datasets as seen in Table \ref{Table2}. The proposed epistemic uncertainty scores appear to be on par with the leading Bayesian benchmark, evenly with out-performance. In MNIST experiments, predictive entropy performs optimally for FashionMNIST, Omniglot and notMNIST, with the epistemic uncertainty scores demonstrating optimal detection abilities for KMNIST. In CIFAR10 experiments, the proposed epistemic uncertainty scores demonstrate significant reduction in FNR95 accross all experiments compared to predictive entropy, and notable improvements in AUROC for SVHN, Places365 and Textures.

\section{Conclusion}

Bayesian neural networks offer principled uncertainty quantification as a possible answer for safeguarding DNNs against erroneous usage. Specifically, epistemic uncertainty scores such a mutual information have been proposed for OoD detection. In this work post-hoc, model-agnostic epistemic uncertainty scores based off measuring disagreement in the maximum logit across posterior samples are proposed and demonstrated improvements over the mutual information baseline for BNNs under mean field variational inference. Generally, this work aims to encourage future investigations on the feature space of BNNs - including but not limited to the logits - to measure epistemic uncertainty for OoD detection. Additionally, this work encourages the evaluation of other forms of posterior inference such as the Laplace approximation \cite{daxberger2021laplace} in conjunction with these proposed scores for OoD detection.


%
%
\bibliographystyle{splncs04}
\bibliography{main}
\end{document}